# *Human-Level Performance on Word Analogy Questions by Latent Relational Analysis*

Peter D. Turney, National Research Council Canada

December 6, 2004

# Human-Level Performance on Word Analogy Questions by Latent Relational Analysis

# Human-Level Performance on Word Analogy Questions by Latent Relational Analysis

## Abstract

This paper introduces Latent Relational Analysis (LRA), a method for measuring relational similarity. LRA has potential applications in many areas, including information extraction, word sense disambiguation, machine translation, and information retrieval. *Relational similarity* is correspondence between relations, in contrast with *attributional similarity*, which is correspondence between attributes. When two words have a high degree of attributional similarity, we call them *synonyms*. When two *pairs* of words have a high degree of relational similarity, we say that their relations are *analogous*. For example, the word pair mason/stone is analogous to the pair carpenter/wood; the relations between mason and stone are highly similar to the relations between carpenter and wood. Past work on semantic similarity measures has mainly been concerned with attributional similarity. For instance, Latent Semantic Analysis (LSA) can measure the degree of similarity between two words, but not between two relations. Recently the Vector Space Model (VSM) of information retrieval has been adapted to the task of measuring relational similarity, achieving a score of 47% on a collection of 374 college-level multiple-choice word analogy questions. In the VSM approach, the relation between a pair of words is characterized by a vector of frequencies of predefined patterns in a large corpus. LRA extends the VSM approach in three ways: (1) the patterns are derived automatically from the corpus (they are not predefined), (2) the Singular Value Decomposition (SVD) is used to smooth the frequency data (it is also used this way in LSA), and (3) automatically generated synonyms are used to explore reformulations of the word pairs. LRA achieves 56% on the 374 analogy questions, statistically equivalent to the average human score of 57%. On the related problem of classifying noun-modifier relations, LRA achieves similar gains over the VSM, while using a smaller corpus.

## 1   Introduction

There are (at least) two kinds of semantic similarity. *Relational similarity* is correspondence between relations, whereas *attributional similarity* is correspondence between attributes (Medin *et al.,* 1990, 1993). Many algorithms have been proposed for measuring the attributional similarity between two words (Lesk, 1969; Church and Hanks, 1989; Ruge, 1992; Dunning, 1993; Smadja, 1993; Resnik, 1995; Landauer and Dumais, 1997; Jiang and Conrath, 1997; Lin, 1998a; Turney, 2001; Budanitsky and Hirst, 2001; Pantel and Lin, 2002; Banerjee and Pedersen, 2003). When two words have a high degree of attributional similarity, we call them *synonyms*. Applications for measures of attributional similarity include the following:

- recognizing synonyms (Landauer and Dumais, 1997; Turney, 2001; Jarmasz and Szpakowicz, 2003; Terra and Clarke, 2003; Turney *et al.,* 2003)

- information retrieval (Deerwester *et al.,* 1990; Dumais, 1993; Hofmann, 1999)

- determining semantic orientation, criticism versus praise (Turney, 2002; Turney and Littman, 2003a)

- grading student essays (Rehder *et al.,* 1998; Foltz *et al.,* 1999)

- measuring textual cohesion (Morris and Hirst, 1991; Barzilay and Elhadad 1997; Foltz *et al.,* 1998; Turney, 2003)

- automatic thesaurus generation (Ruge, 1997; Lin, 1998b; Pantel and Lin, 2002)

- word sense disambiguation (Lesk, 1986; Banerjee and Pedersen, 2003; Patwardhan *et al.*, 2003; Turney, 2004)

Recently, algorithms have been proposed for measuring relational similarity; that is, the semantic similarity between two *pairs* of words (Turney *et al.,* 2003; Turney and Littman, 2003b, 2005). When two word pairs have a high degree of relational similarity, we say they are *analogous*. For example, the pair traffic/street is analogous to the pair water/riverbed. Traffic flows over a street; water flows over a riverbed. A street carries traffic; a riverbed carries water. The relations between traffic and street are similar to the relations between water and riverbed. In fact, this analogy is the basis of several mathematical theories of traffic flow (Daganzo, 1994; Zhang, 2003; Yi *et al.*, 2003). Applications for measures of relational similarity include the following:

- recognizing word analogies (Turney *et al.,* 2003; Turney and Littman, 2003b, 2005)

- classifying semantic relations (Rosario and Hearst, 2001; Rosario *et al.*, 2002; Nastase and Szpakowicz, 2003; Turney and Littman, 2003b, 2005)

- machine translation

- word sense disambiguation

- information extraction (Paice and Black, 2003)

- automatic thesaurus generation (Hearst, 1992a; Berland and Charniak, 1999)

- information retrieval (Hearst, 1992b)

- processing metaphorical text (Martin, 1992; Dolan, 1995)

- identifying semantic roles (Gildea and Jurafsky, 2002)

- analogy-making (Gentner, 1983; Falkenhainer *et al.*, 1989; Falkenhainer, 1990)

Since measures of relational similarity are not as well developed as measures of attributional similarity, the applications of relational similarity (both actual and potential) are not as well known. Section 2 briefly outlines how a measure of relational similarity may be used in the above applications.

We discuss past work on relational similarity and related problems in Section 3. Our work is most closely related to the approach of Turney and Littman (2003b, 2005), which adapts the Vector Space Model (VSM) of information retrieval to the task of measuring relational similarity. In the VSM, documents and queries are represented as vectors, with dimensions corresponding to the words that appear in the given set of documents (Salton and McGill, 1983; Salton and Buckley, 1988; Salton, 1989). The value of an element in a vector is based on the frequency of the corresponding word in the document or query. The similarity between a document and a query is typically measured by the cosine of the angle between the document vector and the query vector (Baeza-Yates and Ribeiro-Neto,1999). Similarly, Turney and Littman (2003b, 2005) represent the semantic relations between a pair of words (e.g., traffic/street) as a vector in which the elements are based on the frequency (in a large corpus) of patterns that connect the pair (e.g., "traffic in the street", "street with traffic"). The relational similarity between two pairs (e.g., traffic/street and water/riverbed) is then measured by the cosine of the angle between their corresponding vectors.

In Section 4, we introduce an algorithm for measuring relational similarity, which we call Latent Relational Analysis (LRA). The algorithm learns from a large corpus of unlabeled, unstructured text, without supervision. LRA extends the VSM approach of Turney and Littman (2003b, 2005) in three ways. Firstly, the connecting patterns are derived automatically from the corpus. Turney and Littman (2003b, 2005), on the other hand, used a fixed set of 64 manually generated patterns (e.g., "… in the …", "… with …"). Secondly, Singular Value Decomposition (SVD) is used to smooth the frequency data. SVD is also part of Latent Semantic Analysis (LSA), which is commonly used for measuring attributional similarity (Deerwester *et al.,* 1990; Dumais, 1993; Landauer and Dumais, 1997). Thirdly, given a word pair such as traffic/street, LRA considers transformations of the word pair, generated by replacing one of the words by synonyms, such as traffic/road, traffic/highway. The synonyms are taken from Lin's (1998b) automatically generated thesaurus.

Measures of attributional similarity have become a popular application for electronic thesauri, such as WordNet and Roget's Thesaurus (Resnik, 1995; Jiang and Conrath, 1997; Budanitsky and Hirst, 2001; Banerjee and Pedersen, 2003; Patwardhan *et al.*, 2003; Jarmasz and Szpakowicz 2003). Recent work suggests that it is possible to use WordNet to measure relational similarity (Veale, 2003). We expect that this will eventually become another common application for electronic thesauri. There is evidence that a hybrid approach, combining corpus-based relational similarity with thesauri-based relational similarity, will prove to be superior to a purebred approach (Turney *et al.,* 2003). However, the scope of the current paper is limited to empirical evaluation of LRA, a purely corpus-based approach.

Section 5 presents our experimental evaluation of LRA with a collection of 374 multiple-choice word analogy questions from the SAT college entrance exam.[1] Table 1 shows a typical SAT question. In the educational testing literature, the first pair (mason:stone) is called the *stem* of the analogy. We use LRA to calculate the relational similarity between the stem and each choice. The choice with the highest relational similarity to the stem is output as the best guess.

Table 1. A sample SAT question.

| Stem: | | mason:stone |
|---|---|---|
| Choices: | (a) | teacher:chalk |
| | (b) | carpenter:wood |
| | (c) | soldier:gun |
| | (d) | photograph:camera |
| | (e) | book:word |
| Solution: | (b) | carpenter:wood |

The average performance of college-bound senior high school students on verbal SAT questions in 2002 (approximately the time when the 374 questions were collected) corresponds to an accuracy of about 57% (Turney and Littman, 2003b, 2005). LRA achieves an accuracy of about 56%. On these same questions, the VSM attained 47%.

---

[1] The College Board has announced that analogies will be eliminated from the SAT in 2005, as part of a shift in the exam to reflect changes in the curriculum.

One application for relational similarity is classifying semantic relations in noun-modifier pairs (Nastase and Szpakowicz, 2003; Turney and Littman, 2003b, 2005). In Section 6, we evaluate the performance of LRA with a set of 600 noun-modifier pairs from Nastase and Szpakowicz (2003). The problem is to classify a noun-modifier pair, such as "laser printer", according to the semantic relation between the head noun (printer) and the modifier (laser). The 600 pairs have been manually labeled with 30 classes of semantic relations. For example, "laser printer" is classified as *instrument*; the printer uses the laser as an instrument for printing. The 30 classes have been manually organized into five general types of semantic relations.

We approach the task of classifying semantic relations in noun-modifier pairs as a supervised learning problem. The 600 pairs are divided into training and testing sets and a testing pair is classified according to the label of its single nearest neighbour in the training set. LRA is used to measure distance (i.e., similarity, nearness). LRA achieves an accuracy of 39.8% on the 30-class problem and 58.0% on the 5-class problem. On the same 600 noun-modifier pairs, the VSM had accuracies of 27.8% (30-class) and 45.7% (Turney and Littman, 2003b, 2005).

We discuss the experimental results, limitations of LRA, and future work in Section 7 and we conclude in Section 8.

## 2  Applications for Measures of Relational Similarity

This section sketches some applications for measures of relational similarity. The following tasks have already been attempted, with varying degrees of success, by researchers who do not explicitly consider relational similarity; only a few of the applications have been framed as involving relational similarity. It is outside the scope of this paper to demonstrate that explicitly considering relational similarity will yield improved performance on all of these various tasks. Our intent in this section is only to suggest that the tasks can be cast in terms of relational similarity.

**Recognizing word analogies:** The problem is, given a stem word pair and a finite list of choice word pairs, select the choice that is most analogous to the stem. This problem was first attempted by a system called Argus (Reitman, 1965), using a small hand-built semantic network. Argus could only solve the limited set of analogy questions that its programmer had anticipated. Argus was based on a spreading activation model and did not explicitly attempt to measure relational similarity. As mentioned in the introduction, Turney and Littman (2003b, 2005) adapt the VSM to measure relational similarity and evaluate the measure using SAT analogy questions. We follow the same approach here. This is discussed in detail in Sections 3 and 5.

**Classifying semantic relations:** The task is to classify the relation between a pair of words. Often the pairs are restricted to noun-modifier pairs, but there are many interesting relations, such as antonymy, that do not occur in noun-modifier pairs. Rosario and Hearst (2001) and Rosario *et al.* (2002) classify noun-modifier relations in the medical domain, using MeSH (Medical Subject Headings) and UMLS (Unified Medical Language System) as lexical resources for representing each noun-modifier pair with a feature vector. They trained a neural network to distinguish 13 classes of semantic relations. Nastase and Szpakowicz (2003) explore a similar approach to classifying general noun-modifier pairs (i.e., not restricted to a particular domain, such as medicine), using WordNet and Roget's Thesaurus as lexical resources. Vanderwende (1994) used hand-built rules, together with a lexical knowledge base, to classify noun-modifier pairs. None of these approaches explicitly involved measuring relational similarity, but any classification of semantic relations necessarily employs some implicit notion of relational similarity, since members of the same class must be relationally similar to some extent. Barker and Szpakowicz (1998) tried a corpus-based approach that explicitly used a measure of relational similarity, but their measure was based on literal matching, which limited its ability to generalize. Turney and Littman (2003b, 2005) used the VSM to measure relational similarity as a component of a single nearest neighbour algorithm. We take the same approach here, substituting LRA for the VSM, in Section 6.

**Machine translation:** Noun-modifier pairs are extremely common in English. For instance, WordNet 2.0 contains more than 26,000 noun-modifier pairs, yet many common noun-modifiers are not in WordNet, especially technical terms. Machine translation cannot rely primarily on manually constructed translation dictionaries for translating noun-modifier pairs, since such dictionaries are necessarily very incomplete. It should be easier to automatically translate noun-modifier pairs when they are first classified by their semantic relations. Consider the pair "electron microscope". Is the semantic relation *purpose* (a microscope for viewing electrons), *instrument* (a microscope that uses electrons), or *material* (a microscope made of electrons)?[2] The answer to this question should facilitate translation of the individual words, "microscope" and "electron", and may also help to determine how the individual words are to be combined in the target language (e.g., what order to put them in, what suffixes to add, what prepositions to add). One possibility would be to reformulate noun-modifier pairs as longer phrases, in the source language, before machine translation. After the semantic relations have been identified, a set of templates can be used to reformulate the noun-modifier pairs. For example, "If *'A B'* is classified as *instrument*, reformulate it as *'B that uses A'*."

**Word sense disambiguation:** Noun-modifier pairs are almost always monosemous (Yarowsky, 1993, 1995). We hypothesize that the implicit semantic relation between the two words in the pair narrowly constrains the possible senses of the words. More generally, we conjecture that the intended sense of a word is determined by its semantic relations with the other words in the surrounding text. If we can identify the semantic relations between the given word and its context, then we can disambiguate the given word. Consider the noun-modifier pair "plant food". In isolation, "plant" could refer to an industrial plant or a living organism. Once we have determined that the implicit semantic relation in "plant food" is *beneficiary* (the plant benefits from the food), as opposed to, say, *location at* (the food is located at the plant), the sense of "plant" is constrained to "living organism".[3] As far as we know, no existing word sense disambiguation system makes explicit use of a measure of relational similarity, but we believe this is a promising approach to word sense disambiguation.

**Information extraction:** Paice and Black (2003) describe a system that identifies the key terms in a document and then discovers the semantic relations between the key terms. For example, from a document entitled *Soil Temperature and Water Content, Seeding Depth, and Simulated Rainfall Effects on Winter Wheat Emergence*, they were able to automatically extract the terms "seeding depth", "emergence", and "winter wheat", and then discover the semantic relations among these three terms. Their corpus consisted of papers on agriculture (crop husbandry) and their algorithm is limited to relations among *influences* (e.g., seeding depth), *properties* (e.g., emergence), and

---

[2] These three semantic relations are from the 30 classes of Nastase and Szpakowicz (2003).

[3] These two semantic relations are from the 30 classes of Nastase and Szpakowicz (2003).

*objects* (e.g., winter wheat). They used manually generated rules to identify the semantic relations, but an alternative approach would be to use supervised learning, which would make it easier to expand the system to a wider range of documents and a larger set of semantic relations. For example, a measure of relational similarity could be used as a distance measure in a nearest neighbour algorithm (Turney and Littman, 2003b, 2005).

**Automatic thesaurus generation:** Hearst (1992a) presents an algorithm for learning hyponym *(type of)* relations from a corpus and Berland and Charniak (1999) describe how to learn meronym *(part of)* relations from a corpus. These algorithms could be used to automatically generate a thesaurus or dictionary, but we would like to handle more relations than hyponymy and meronymy. WordNet distinguishes more than a dozen semantic relations between words (Fellbaum, 1998) and Nastase and Szpakowicz (2003) list 30 semantic relations for noun-modifier pairs. Hearst (1992a) and Berland and Charniak (1999) use manually generated rules to mine text for semantic relations. Turney and Littman (2003b, 2005) also use a manually generated set of 64 patterns. LRA does not use a predefined set of patterns; it learns patterns from a large corpus. Instead of manually generating new rules or patterns for each new semantic relation, it is possible to automatically learn a measure of relational similarity that can handle arbitrary semantic relations. A nearest neighbour algorithm can then use this relational similarity measure to learn to classify according to any given set of classes of relations.

**Information retrieval:** Current search engines are based on attributional similarity; the similarity of a query to a document depends on correspondence between the attributes of the query and the attributes of the documents (Baeza-Yates and Ribeiro-Neto,1999). Typically the correspondence is exact matching of words or root words. Latent Semantic Indexing allows more flexible matching, but it is still based on attributional similarity (Deerwester *et al.,* 1990; Dumais, 1993). Hearst (1992b) outlines an algorithm for recognizing text in which an agent *enables* or *blocks* an event. The algorithm was designed for use in information retrieval applications, but it has not yet been implemented. If we could reliably classify semantic relations, then we could ask new kinds of search queries:

- find all documents about things that have been *enabled* by the Canadian government
- find all documents about things that have a *causal* relation with cancer
- find all documents about things that have an *instrument* relation with printing

Existing search engines cannot recognize the implicit *instrument* relation in "laser printer", so the query "instrument and printing" will miss many relevant documents. A measure of relational similarity could be used as a component in a supervised learning system that learns to identify semantic relations between words in documents. These semantic relations could then be added to the index of a conventional (attributional) search engine. Alternatively, a search engine could compare a query to a document using a similarity measure that takes into account both relational similarity and attributional similarity. A query might be phrased as a word analogy problem:

- find all documents about things that are to printers as lasers are to printers
- find all documents about things that are to dogs as catnip is to cats
- find all documents about things that are to Windows as grep is to Unix

**Processing metaphorical text:** Metaphorical language is very common in our daily life; so common that we are usually unaware of it (Lakoff and Johnson, 1980; Lakoff, 1987).

Martin (1992) notes that even technical dialogue, such as computer users asking for help, is often metaphorical:

- How can I *kill* a process?

- How can I *get into* the LISP interpreter?

- Tell me how to *get out of* Emacs.

Human-computer dialogue systems are currently limited to very simple, literal language. We believe that the task of mapping metaphorical language to more literal language can be approached as a kind of word analogy problem:

- *kill* is to an *organism* as *stop* is to a *process*

- *get into* is to a *container* as *start* is to the *LISP interpreter*

- *get out of* is to a *container* as *stop* is to the *Emacs editor*

We conjecture that a measure of relational similarity can be used to solve these kinds of word analogy problems, and thus facilitate computer processing of metaphorical text. Gentner *et al.* (2001) argue that novel metaphors are understood using analogy, but conventional metaphors are simply recalled from memory. A *conventional* metaphor is a metaphor that has become entrenched in our language (Lakoff and Johnson, 1980). Dolan (1995) describes an algorithm that can recognize conventional metaphors, but is not suited to novel metaphors. This suggests that it may be fruitful to combine Dolan's (1995) algorithm for handling conventional metaphorical language with an algorithm such as LRA for handling novel metaphors.

**Identifying semantic roles:** A semantic frame for an event such as *judgement* contains semantic roles such as *judge*, *evaluee*, and *reason*, whereas an event such as *statement* contains roles such as *speaker*, *addressee*, and *message* (Gildea and Jurafsky, 2002). The task of identifying semantic roles is to label the parts of a sentence according to their semantic roles. For example (Gildea and Jurafsky, 2002):

- [*judge:* She] **blames** [*evaluee:* the Government] [*reason:* for failing to do enough to help].

- [*message:* "I'll knock on your door at quarter to six"] [*speaker:* Susan] **said**.

We believe that it may be helpful to view semantic frames and their semantic roles as sets of semantic relations. For example, the frame *judgement(judge, evaluee, reason)* can be viewed as three pairs, *<judgement, judge>*, *<judgement, evaluee>*, and *<judgement, reason>*. This is the assumption behind Dependency Grammar, that syntactic structure consists of lexical nodes (representing words) and binary relations (dependencies) linking them (Mel'cuk, 1988). Thus a measure of relational similarity should help us to identify semantic roles. The sentence could first be parsed using Dependency Grammar (Lin, 1998c), and then the binary relations could be classified using supervised learning (e.g., a nearest neighbour algorithm with a distance measure based on relational similarity).

**Analogy-making:** French (2002) cites Structure Mapping Theory (SMT) (Gentner, 1983) and its implementation in the Structure Mapping Engine (SME) (Falkenhainer *et al.*, 1989) as the most influential work on modeling of analogy-making. The goal of computational modeling of analogy-making is to understand how people form complex, structured analogies. SME takes representations of a source domain and a target domain, and produces an analogical mapping between the source and target. The domains are given structured propositional representations, using predicate logic. These descriptions include attributes, relations, and higher-order relations (expressing relations between relations). The analogical mapping connects source domain relations to target domain relations. Each individual connection in an analogical mapping implies that the connected relations are similar; thus, SMT requires a measure of relational similarity, in order to form maps. Early versions of SME only mapped identical relations, but later versions of SME allowed similar, non-identical relations to match (Falkenhainer, 1990). However, the focus of research in analogy-making has been on the mapping process as a whole, rather than measuring the similarity between any two particular relations, hence the similarity measures used in SME at the level of individual connections are somewhat rudimentary. We believe that a more sophisticated measure of relational similarity, such as LRA, may enhance the performance of SME. Likewise, the focus of our work here is on the similarity between particular relations, and we ignore systematic mapping between sets of relations, so our work may also be enhanced by integration with SME.

For many of the above examples, it not yet known whether there is a benefit to framing the problems in terms of relational similarity, but the list illustrates the wide range of potential applications.

## 3 Related Work

Let $R_1$ be the semantic relation (or set of relations) between a pair of words, *A* and *B*, and let $R_2$ be the semantic relation (or set of relations) between another pair, *C* and *D*. We wish to measure the relational similarity between $R_1$ and $R_2$. The relations $R_1$ and $R_2$ are not given to us; our task is to infer these hidden (latent) relations and then compare them.

Latent Relational Analysis builds on the Vector Space Model of Turney and Littman (2003b, 2005). In the VSM approach to relational similarity, we create vectors, $r_1$ and $r_2$, that represent features of $R_1$ and $R_2$, and then measure the similarity of $R_1$ and $R_2$ by the cosine of the angle $\theta$ between $r_1$ and $r_2$:

$$r_1 = \left\langle r_{1,1}, \ldots, r_{1,n} \right\rangle$$

$$r_2 = \left\langle r_{2,1}, \ldots r_{2,n} \right\rangle$$

$$\mathrm{cosine}(\theta) = \frac{\sum_{i=1}^{n} r_{1,i}\, r_{2,i}}{\sqrt{\sum_{i=1}^{n} \left(r_{1,i}\right)^2 \cdot \sum_{i=1}^{n} \left(r_{2,i}\right)^2}} = \frac{r_1 \cdot r_2}{\sqrt{r_1 \cdot r_1} \cdot \sqrt{r_2 \cdot r_2}} = \frac{r_1 \cdot r_2}{\|r_1\| \cdot \|r_2\|}\,.$$

We create a vector, *r*, to characterize the relationship between two words, *X* and *Y*, by counting the frequencies of various short phrases containing *X* and *Y*. Turney and Littman (2003b, 2005) use a list of 64 joining terms, such as "of", "for", and "to", to form 128 phrases that contain *X* and *Y*, such as *"X of Y", "Y of X", "X for Y", "Y for X", "X to Y",* and *"Y to X"*. These phrases are then used as queries for a search engine and the number of hits (matching documents) is recorded for each query. This process yields a vector of 128 numbers. If the number of hits for a query is *x*, then the corresponding element in the vector *r* is log(*x*+1). Several authors report that the logarithmic transformation of frequencies improves cosine-based similarity measures (Salton and Buckley, 1988; Ruge, 1992; Lin, 1998a).

Turney and Littman (2003b, 2005) evaluated the VSM approach by its performance on 374 SAT analogy questions, achieving a score of 47%. Since there are five choices for each question, the expected score for random guessing is 20%. To answer a multiple-choice analogy question, vectors are created for the stem pair and each choice pair, and then cosines are calculated for the angles between the stem pair and each choice pair. The best guess is the choice pair with the highest cosine. We use the same set of analogy questions to evaluate LRA in Section 5.

The VSM was also evaluated by its performance as a distance (nearness) measure in a supervised nearest neighbour classifier for noun-modifier semantic relations (Turney and Littman, 2003b, 2005). The evaluation used 600 hand-labeled noun-modifier pairs from Nastase and Szpakowicz (2003). An testing pair is classified by searching for its single nearest neighbour in the labeled training data. The best guess is the label for the training pair with the highest cosine. LRA is evaluated with the same set of noun-modifier pairs in Section 6.

Turney and Littman (2003b, 2005) used the AltaVista search engine to obtain the frequency information required to build vectors for the VSM. Thus their corpus was the set of all web pages indexed by AltaVista. At the time, the English subset of this corpus consisted of about $5 \times 10^{11}$ words. Around April 2004, AltaVista made substantial changes to their search engine, removing their advanced search operators. Their search engine no longer supports the asterisk operator, which was used by Turney and Littman (2003b, 2005) for stemming and wild-card searching. AltaVista also changed their policy towards automated searching, which is now forbidden.[4]

Turney and Littman (2003b, 2005) used AltaVista's hit count, which is the number of *documents* (web pages) matching a given query, but LRA uses the number of *passages* (strings) matching a query. In our experiments with LRA (Sections 5 and 6), we use a local copy of the Waterloo MultiText System (Clarke *et al.*, 1998; Terra and Clarke, 2003), running on a 16 CPU Beowulf Cluster, with a corpus of about $5 \times 10^{10}$ English words. The Waterloo MultiText System (WMTS) is a distributed (multiprocessor) search engine, designed primarily for passage retrieval (although document retrieval is possible, as a special case of passage retrieval). The text and index require approximately one terabyte of disk space. Although AltaVista only gives a rough estimate of the number of matching documents, the Waterloo MultiText System gives exact counts of the number of matching passages.

Turney *et al.* (2003) combine 13 independent modules to answer SAT questions. The final output of the system was based on a weighted combination of the outputs of each individual module. The best of the 13 modules was the VSM, described above. Although each of the individual component modules did not require training data, the module combination algorithm was supervised, so the SAT questions were divided into training and testing sets. The training questions were used to tune the combination weights. We compare the performance of this approach to LRA in Section 5.

The VSM was first developed for information retrieval (Salton and McGill, 1983; Salton and Buckley, 1988; Salton, 1989) and it is at the core of most modern search engines (Baeza-Yates and Ribeiro-Neto,1999). In the VSM approach to information retrieval, queries and documents are represented by vectors. Elements in these vectors are based on the frequencies of words in the corresponding queries and documents. The frequencies are usually transformed by various formulas and weights, tailored to improve the effectiveness of the search engine (Salton, 1989). The similarity between a query and a document is measured by the cosine of the angle between their corresponding vectors. For a given query, the search engine sorts the matching documents in order of decreasing cosine.

---

[4] See http://www.altavista.com/robots.txt for AltaVista's current policy towards "robots" (software for automatically gathering web pages or issuing batches of queries). The protocol of the "robots.txt" file is explained in http://www.robotstxt.org/wc/robots.html.

The VSM approach has also been used to measure the similarity of words (Lesk, 1969; Ruge, 1992; Pantel and Lin, 2002). Pantel and Lin (2002) clustered words according to their similarity, as measured by a VSM. Their algorithm is able to discover the different senses of a word, using unsupervised learning.

Document-query similarity (Salton and McGill, 1983; Salton and Buckley, 1988) and word similarity (Lesk, 1969; Ruge, 1992; Pantel and Lin, 2002) are both instances of attributional similarity. Turney *et al.* (2003) and Turney and Littman (2003b, 2005) show that the VSM is also applicable to relational similarity.

Latent Semantic Analysis extends the VSM approach to information retrieval by using the Singular Value Decomposition to smooth the vectors, which helps to handle noise and sparseness in the data (Deerwester *et al.,* 1990; Dumais, 1993; Landauer and Dumais, 1997). The SVD improves both document-query similarity measures (Deerwester *et al.,* 1990; Dumais, 1993) and word similarity measures (Landauer and Dumais, 1997). LRA also uses SVD to smooth vectors, as we discuss in the next section.

# 4  Latent Relational Analysis

LRA takes as input a set of word pairs and produces as output a measure of the relational similarity between any two of the input pairs. In our experiments, the input set contains from 600 to 2,244 word pairs. If there are not enough input pairs, SVD will not be effective; if there are too many pairs, SVD will not be efficient. The similarity measure is based on cosines, so the degree of similarity can range from -1 (dissimilar; $\theta = 180°$) to +1 (similar; $\theta = 0°$). Before applying SVD, the vectors are completely nonnegative, which implies that the cosine can only range from 0 to +1, but SVD introduces negative values, so it is possible for the cosine to be negative, although we have never observed this in our experiments.

LRA relies on three resources, (1) a search engine with a very large corpus of text, (2) a broad-coverage thesaurus of synonyms, and (3) an efficient implementation of SVD. LRA does not use labeled data, structured data, or supervised learning.

In the following experiments, we use a local copy of the Waterloo MultiText System (Clarke *et al.*, 1998; Terra and Clarke, 2003).[5] The corpus consists of about $5 \times 10^{10}$ English words, gathered by a web crawler, mainly from US academic web sites. The WMTS is well suited to LRA, because it scales well to large corpora (one terabyte, in our case), it gives exact frequency counts (unlike most web search engines), it is designed for passage retrieval (rather than document retrieval), and it has a powerful query syntax.

As a source of synonyms, we use Lin's (1998b) automatically generated thesaurus. This thesaurus is available through an online interactive demonstration or it can be downloaded.[6] We used the online demonstration, since the downloadable version seems to be less complete. For each word in the input set of word pairs, we automatically query the online demonstration and fetch the resulting list of synonyms. As a courtesy, we insert a 20 second delay between each query.

---

[5] See http://multitext.uwaterloo.ca/.

Lin's thesaurus was generated by parsing a corpus of about $5 \times 10^7$ English words, consisting of text from the Wall Street Journal, San Jose Mercury, and AP Newswire (Lin, 1998b). The parser was used to extract pairs of words and their grammatical relations, according to Dependency Grammar (Mel'cuk, 1988). Words were then clustered into synonym sets, based on the similarity of their grammatical relations. Two words were judged to be highly similar when they tended to have the same kinds of grammatical relations with the same sets of words. Given a word and its part of speech, Lin's thesaurus provides a list of words, sorted in order of decreasing attributional similarity. This sorting is convenient for LRA, since it makes it possible to focus on words with higher attributional similarity and ignore the rest. WordNet, in contrast, given a word and its part of speech, provides a list of words grouped by the possible senses of the given word, with groups sorted by the frequencies of the senses. WordNet's sorting does not directly correspond to sorting by degree of attributional similarity, although various algorithms have been proposed for deriving attributional similarity from WordNet (Resnik, 1995; Jiang and Conrath, 1997; Budanitsky and Hirst, 2001; Banerjee and Pedersen, 2003).

We use Rohde's SVDLIBC implementation of the Singular Value Decomposition, which is based on SVDPACKC (Berry, 1992).[7] SVD decomposes a matrix $\mathbf{X}$ into a product of three matrices $\mathbf{U}\Sigma\mathbf{V}^T$, where $\mathbf{U}$ and $\mathbf{V}$ are in column orthonormal form (i.e., the columns are orthogonal and have unit length: $\mathbf{U}^T\mathbf{U} = \mathbf{V}^T\mathbf{V} = \mathbf{I}$) and $\Sigma$ is a diagonal matrix of *singular values* (hence SVD) (Golub and Van Loan, 1996). If $\mathbf{X}$ is of rank $r$, then $\Sigma$ is also of rank $r$. Let $\Sigma_k$, where $k < r$, be the diagonal matrix formed from the top $k$ singular values, and let $\mathbf{U}_k$ and $\mathbf{V}_k$ be the matrices produced by selecting the corresponding columns from $\mathbf{U}$ and $\mathbf{V}$. The matrix $\mathbf{U}_k\Sigma_k\mathbf{V}_k^T$ is the matrix of rank $k$ that best approximates the original matrix $\mathbf{X}$, in the sense that it minimizes the approximation errors. That is, $\hat{\mathbf{X}} = \mathbf{U}_k\Sigma_k\mathbf{V}_k^T$ minimizes $\left\|\hat{\mathbf{X}} - \mathbf{X}\right\|_F$ over all matrices $\hat{\mathbf{X}}$ of rank $k$, where $\left\|...\right\|_F$ denotes the Frobenius norm (Golub and Van Loan, 1996). We may think of this matrix $\mathbf{U}_k\Sigma_k\mathbf{V}_k^T$ as a "smoothed" or "compressed" version of the original matrix. In LRA, SVD is used to reduce noise and compensate for sparseness.

Briefly, LRA proceeds as follows. First take the input set of word pairs and expand the set by substituting synonyms for each of the words in the word pairs. Then, for each word pair, search in the corpus for all phrases (up to a fixed maximum length) that begin with one member of the pair and end with the other member of the pair. Build a set of patterns by examining these phrases. Ignore the initial and final words (which come from the input set) and focus on the intervening words. Make a list of all patterns of intervening words and, for each pattern, count the number of word pairs (in the expanded set) that occur in phrases that match the given pattern. The pattern list will typically contain millions of patterns. Take the top few thousand most frequent patterns and drop the rest. Build a matrix $\mathbf{X}$ in which the rows correspond to the word pairs and the columns correspond to the selected higher-frequency patterns. The value of a cell in the matrix is based on the number of times the corresponding word pair appears in a phrase that matches the corresponding pattern. Apply SVD to the matrix and approximate the original matrix with $\mathbf{U}_k \Sigma_k \mathbf{V}_k^T$. Now we are ready to calculate relational similarity. Suppose we wish to compare a word pair *A:B* to a word pair *C:D*. Look for the row that corresponds to *A:B* and the row that corresponds to *C:D* and calculate the cosine of the two row vectors. Let's call this cosine the *original cosine*. For every *A':B'* that was generated by substituting synonyms in *A:B* and for every *C':D'* that was generated by substituting synonyms in *C:D*, calculate the corresponding cosines, which we will call the *reformulated cosines*. Finally, the relational similarity between *A:B* and *C:D* is the average of all cosines (original and reformulated) that are greater than or equal to the original cosine.

---

[6] The online demonstration is at http://www.cs.ualberta.ca/~lindek/demos/depsim.htm and the downloadable version is at http://armena.cs.ualberta.ca/lindek/downloads/sims.lsp.gz.

[7] SVDLIBC is available at http://tedlab.mit.edu/~dr/SVDLIBC/ and SVDPACKC is available at http://www.netlib.org/svdpack/.

This sketch of LRA omits a few significant points. We will go through each step in more detail, using an example to illustrate the steps. Table 2 gives an example of a SAT question (Claman, 2000). Let's suppose that we wish to calculate the relational similarity between the pair quart:volume and the pair mile:distance.

Table 2. A sample SAT question.

| Stem: | | quart:volume |
|-------|-----|--------------|
| Choices: | (a) | day:night |
| | (b) | mile:distance |
| | (c) | decade:century |
| | (d) | friction:heat |
| | (e) | part:whole |
| Solution: | (b) | mile:distance |

1. **Find alternates:** Replace each word in each input pair with similar words. Look in Lin's (1998b) thesaurus for the top ten most similar words for each word in each pair. For a given word, Lin's thesaurus has lists of similar words for each possible part of speech, but we assume that the part-of-speech of the given word is unknown. Lin's thesaurus gives a numerical score for the attributional similarity of each word, compared to the given word, so we can merge the lists for each part of speech (when the given word has more than one possible part of speech) and sort the merged list by the numerical scores. When a word appears in two or more lists, only enter it once in the merged list, using its maximum score. For a given word pair, only substitute similar words for one member of the pair at a time (this constraint limits the divergence in meaning). Avoid similar words that seem unusual in any way (e.g., hyphenated words, words with three characters or less, words with non-alphabetical characters, multi-word phrases, and capitalized words). The first column in Table 3 shows the alternate pairs that are generated for the original pair quart:volume.

2. **Filter alternates:** Find how often each pair (originals and alternates) appears in the corpus. Send a query to the WMTS to count the number of sequences of five consecutive words, such that one of the five words matches the first member of the pair and another of the five matches the second member of the pair (the order does not matter). For the pair quart:volume, the WMTS query is simply '[5]>("quart"^"volume")'. Sort the alternate pairs by their frequency. Take the top three alternate pairs and reject the rest. Keep the original pair. The last column in Table 3 shows the pairs that are selected.

Table 3. Alternate forms of the original pair quart:volume.

| Word pair | Lin's similarity score for the alternate word compared to the original | Frequency of pair in corpus | Filtering step |
|---|---|---|---|
| quart:volume | NA | 632 | accept (original pair) |
| pint:volume | 0.209843 | 372 | |
| gallon:volume | 0.158739 | 1500 | accept (top alternate) |
| liter:volume | 0.122297 | 3323 | accept (top alternate) |
| squirt:volume | 0.0842603 | 54 | |
| pail:volume | 0.083708 | 28 | |
| vial:volume | 0.083708 | 373 | |
| pumping:volume | 0.0734792 | 1386 | accept (top alternate) |
| ounce:volume | 0.0709759 | 430 | |
| spoonful:volume | 0.0704245 | 42 | |
| tablespoon:volume | 0.0685988 | 96 | |
| quart:turnover | 0.228795 | 0 | |
| quart:output | 0.224934 | 34 | |
| quart:export | 0.206013 | 7 | |
| quart:value | 0.203389 | 266 | |
| quart:import | 0.185549 | 16 | |
| quart:revenue | 0.184562 | 0 | |
| quart:sale | 0.16854 | 119 | |
| quart:investment | 0.160734 | 11 | |
| quart:earnings | 0.156212 | 0 | |
| quart:profit | 0.155507 | 24 | |

3. **Find phrases:** For each pair (originals and alternates), make a list of phrases in the corpus that contain the pair. Query the WMTS for all phrases that begin with one member of the pair, end with the other member of the pair, and have one to three intervening words (i.e., a total of three to five words). We wish to ignore suffixes, so that a phrase such as "volume measured in quarts" will count as a match for quart:volume. The WMTS query syntax has an asterisk operator that matches suffixes (e.g., "quart*" will match both "quart" and "quarts"), but this operator is computationally intensive. Instead of using the asterisk operator, we use a few simple heuristics to add (or subtract) some of the most common suffixes to (or from) each word (e.g., "s", "ing", "ed"). For each pair, several queries are sent to the WMTS, specifying different suffixes (e.g., "quart" and "quarts"), different orders (e.g., phrases that start with "quart" and end with "volume", and vice versa), and different numbers of intervening words (1, 2, or 3). Note that a nonsensical suffix does no harm; the query simply returns no matching phrases. This step is relatively easy with the WMTS, since it is designed for passage retrieval. With a document retrieval search engine, it would be necessary to fetch each document that matches a query, and then scan through the document, looking for the phrase that matches the query. The WMTS simply returns a list of the phrases that match the query. Table 4 gives some examples of phrases in the corpus that match the pair quart:volume.

Table 4 Phrases that contain quart:volume.

| | |
|---|---|
| quarts liquid volume | volume in quarts |
| quarts of volume | volume capacity quarts |
| quarts in volume | volume being about two quarts |
| quart total volume | volume of milk in quarts |
| quart of spray volume | volume include measures like quart |

4. **Find patterns:** For each phrase that is found in the previous step, build patterns from the intervening words. A pattern is constructed by replacing any or all or none of the intervening words with a wild card. For example, the phrase "quart of spray volume" contains the intervening sequence "of spray", which yields four patterns, "of spray", "of spray", "* spray", "of *", and "* *", where "*" is the wild card. This step can produce millions of patterns, given a couple of thousand input pairs. Keep the top 4,000 most frequent patterns and throw away the rest. To count the frequencies of the patterns, scan through the lists of phrases sequentially, generate patterns from each phrase, and put the patterns in a database.[8] The patterns are key fields in the database and the frequencies are the corresponding value fields. The first time a pattern is encountered, its frequency is set to one. The frequency is incremented each time the pattern is encountered with a new pair. After all of the phrases have been processed, the top 4,000 patterns can easily be extracted from the database.

5. **Map pairs to rows:** In preparation for building the matrix $\mathbf{X}$, create a mapping of word pairs to row numbers. Assign a row number to each word pair (originals and alternates), unless a word pair has no corresponding phrases (from step 3). If a word pair has no phrases, the corresponding row vector would be a zero vector, so there is no point to including it in the matrix. For each word pair *A:B*, create a row for *A:B* and another row for *B:A*. This will make the matrix more symmetrical, reflecting our knowledge that the relational similarity between *A:B* and *C:D* should be the same as the relational similarity between *B:A* and *D:C*. The intent is to assist SVD by enforcing this symmetry in the matrix.

6. **Map patterns to columns:** Create a mapping of the top 4,000 patterns to column numbers. Since the patterns have been selected for their high frequency, it is not likely that any of the column vectors will be zero vectors (unless the input set of pairs is small or the words are very rare). For each pattern, *P*, create a column for "*word_1 P word_2*" and another column for "*word_2 P word_1*". For example, the frequency of "quarts of volume" will be distinguished from the frequency of "volume of quarts". Thus there will be 8,000 columns.

7. **Generate a sparse matrix:** Generate the matrix $\mathbf{X}$ in sparse matrix format, suitable for input to SVDLIBC. The value for the cell in row *i* and column *j* is the frequency of the *j*-th pattern (see step 6) in phrases that contain the *i*-th word pair (see step 5). We do not issue queries to the WMTS to obtain these frequencies, since we found it faster to sequentially scan the lists of phrases from step 3, counting the number of pattern matches. Table 5 gives some examples of pattern frequencies for quart:volume. A pattern of *N* tokens (words or wild cards) can only match phrases in which there are exactly *N* intervening words (i.e., phrases in which there are a total of *N*+2 words; *N* = 1, 2, 3). A wild card can match any word, but one wild card can only match one word.

---

[8] We use Perl with the Berkeley DB package. See http://www.sleepycat.com/.

Table 5. Frequencies of various patterns for quart:volume

| | Frequency | |
|---|---|---|
| | $word_1 =$ quart, quarts | |
| | $word_2 =$ volume, volumes | |
| $P$ | "$word_1$ $P$ $word_2$" | "$word_2$ $P$ $word_1$" |
| "in" | 4 | 10 |
| "* of" | 1 | 0 |
| "of *" | 5 | 2 |
| "* *" | 19 | 16 |

8. **Calculate entropy:** Apply log and entropy transformations to the sparse matrix (Landauer and Dumais, 1997). These transformations have been found to be very helpful for information retrieval (Harman, 1986; Dumais, 1990). Let $x_{i,j}$ be the cell in row $i$ and column $j$ of the matrix **X** from step 7. Let $m$ be the number of rows in **X** and let $n$ be the number of columns. We wish to weight the cell $x_{i,j}$ by the entropy of the $j$-th column. To calculate the entropy of the column, we need to convert the column into a vector of probabilities. Let $p_{i,j}$ be the probability of $x_{i,j}$, calculated by normalizing the column vector so that the sum of the elements is one, $p_{i,j} = x_{i,j} \Big/ \sum_{k=1}^{m} x_{k,j}$. The entropy of the $j$-th column is then $H_j = -\sum_{k=1}^{m} p_{k,j} \log(p_{k,j})$. Entropy is at its maximum when $p_{i,j}$ is a uniform distribution, $p_{i,j} = 1/m$, in which case $H_j = \log(m)$. Entropy is at its minimum when $p_{i,j}$ is 1 for some value of $i$ and 0 for all other values of $i$, in which case $H_j = 0$. We want to give more weight to columns (patterns) with frequencies that vary substantially from one row (word pair) to the next, and less weight to columns that are uniform. Therefore we weight the cell $x_{i,j}$ by $w_j = 1 - H_j / \log(m)$, which varies from 0 when $p_{i,j}$ is uniform to 1 when entropy is minimal. We also apply the log transformation to frequencies, $\log(x_{i,j} + 1)$. For all $i$ and all $j$, replace the original value $x_{i,j}$ in **X** by the new value $w_j \log(x_{i,j} + 1)$. This is similar to the TF-IDF transformation (Term Frequency-Inverse Document Frequency) that is familiar in information retrieval (Salton and Buckley, 1988; Baeza-Yates and Ribeiro-Neto,1999).

9. **Apply SVD:** After the log and entropy transformations have been applied to the matrix **X**, run SVDLIBC. This produces the matrices **U**, $\Sigma$, and **V**, where $\mathbf{X} = \mathbf{U}\Sigma\mathbf{V}^T$. In the subsequent steps, we will be calculating cosines for row vectors. For this purpose, we can simplify calculations by dropping **V**. The cosine of two vectors is their dot product, after they have been normalized to unit length. The matrix $\mathbf{X}\mathbf{X}^T$ contains the dot products of all of the row vectors. We can find the dot product of the $i$-th and $j$-th row vectors by looking at the cell in row $i$, column $j$ of the matrix $\mathbf{X}\mathbf{X}^T$. Since $\mathbf{V}^T\mathbf{V} = \mathbf{I}$, we have $\mathbf{X}\mathbf{X}^T = \mathbf{U}\Sigma\mathbf{V}^T(\mathbf{U}\Sigma\mathbf{V}^T)^T = \mathbf{U}\Sigma\mathbf{V}^T\mathbf{V}\Sigma\mathbf{U}^T = \mathbf{U}\Sigma(\mathbf{U}\Sigma)^T$, which means that we can calculate cosines with the smaller matrix $\mathbf{U}\Sigma$, instead of using $\mathbf{X} = \mathbf{U}\Sigma\mathbf{V}^T$ (Deerwester *et al.,* 1990).

10. **Projection:** Calculate $\mathbf{U}_k\Sigma_k$, where $k = 300$. The matrix $\mathbf{U}_k\Sigma_k$ has the same number of rows as **X**, but only 300 columns (compared with 8,000; see step 6). We can compare two word pairs by calculating the cosine of the corresponding row vectors in $\mathbf{U}_k \Sigma_k$. The row vector for each word pair has been projected from the original 8,000 dimensional space into a new 300 dimensional space. The value $k = 300$ is recommended by Landauer and Dumais (1997) for measuring the attributional similarity between words. We investigate other values in Section 5.

11. **Evaluate alternates:** Let *A:B* and *C:D* be two word pairs in the input set. Assume that we want to measure their relational similarity. From step 2, we have four versions of *A:B,* the original pair and three alternate pairs. Likewise, we have four versions of *C:D*. Therefore we have sixteen ways to compare a version of *A:B* with a version of *C:D*. Look for the row vectors in $\mathbf{U}_k \Sigma_k$ that correspond to the four versions of *A:B* and the four versions of *C:D* and calculate the sixteen cosines. For example, suppose *A:B* is quart:volume and *C:D* is mile:distance. Table 6 gives the cosines for the sixteen combinations. *A:B::C:D* expresses the analogy *"A is to B as C is to D".*

Table 6. The sixteen combinations and their cosines.

| Word pairs | Cosine | Cosine ≥ original pairs |
|---|---|---|
| quart:volume::mile:distance | 0.524568 | yes (original pairs) |
| quart:volume::feet:distance | 0.463552 | |
| quart:volume::mile:length | 0.634493 | yes |
| quart:volume::length:distance | 0.498858 | |
| liter:volume::mile:distance | 0.735634 | yes |
| liter:volume::feet:distance | 0.686983 | yes |
| liter:volume::mile:length | 0.744999 | yes |
| liter:volume::length:distance | 0.576477 | yes |
| gallon:volume::mile:distance | 0.763385 | yes |
| gallon:volume::feet:distance | 0.709965 | yes |
| gallon:volume::mile:length | 0.781394 | yes (highest cosine) |
| gallon:volume::length:distance | 0.614685 | yes |
| pumping:volume::mile:distance | 0.411644 | |
| pumping:volume::feet:distance | 0.439250 | |
| pumping:volume::mile:length | 0.446202 | |
| pumping:volume::length:distance | 0.490511 | |

12. **Calculate relational similarity:** The relational similarity between *A:B* and *C:D* is the average of the cosines, among the sixteen cosines from step 11, that are greater than or equal to the cosine of the original pairs. For quart:volume and mile:distance, the third column in Table 6 shows which reformulations are used to calculate the average. For these two pairs, the average of the selected cosines is 0.677258. Table 7 gives the cosines for the sample SAT question, introduced in Table 2. The choice pair with the highest average cosine (column three in the table) is the solution for this question; LRA answers the question correctly. For comparison, column four gives the cosines for the original pairs and column five gives the highest cosine (the maximum over the sixteen reformulations). For this particular SAT question, there is one choice that has the highest cosine for all three columns (choice (b)), although this is not true in general. However, note that the gap between the first choice (b) and the second choice (d) is largest for the average cosines.

Table 7. Cosines for the sample SAT question.

| Stem: | | quart:volume | Average cosines | Original cosines | Highest cosines |
|---|---|---|---|---|---|
| Choices: | (a) | day:night | 0.373725 | 0.326610 | 0.443079 |
| | (b) | mile:distance | **0.677258** | 0.524568 | 0.781394 |
| | (c) | decade:century | 0.388504 | 0.327201 | 0.469610 |
| | (d) | friction:heat | 0.427860 | 0.336138 | 0.551676 |
| | (e) | part:whole | 0.370172 | 0.329997 | 0.408357 |
| Solution: | (b) | mile:distance | 0.677258 | 0.524568 | 0.781394 |
| Gap: | (b)-(d) | | 0.249398 | 0.188430 | 0.229718 |

Steps 11 and 12 can be repeated for each two input pairs that are to be compared.

Table 8 lists the numerical parameters in LRA and the step in which each parameter appears. The value of *max_phrase* is set to *max_inter* + 2 and the value of *num_combinations* is (*num_filter* + 1)$^2$. The remaining parameters have been set to arbitrary values (see Section 7).

Table 8. Numerical parameters in LRA.

| Step number in LRA | Description of parameter | Default value of parameter | Parameter name |
|---|---|---|---|
| Step 1 | the number of similar words that are considered as alternates for each input word | 10 | *num_sim* |
| Step 2 | the maximum phrase length | 5 | *max_phrase* |
| Step 2 | the number of alternate word pairs that pass through the filter (not counting the original pair) | 3 | *num_filter* |
| Step 3 | the minimum number of intervening words in a phrase | 1 | *min_inter* |
| Step 3 | the maximum number of intervening words in a phrase | 3 | *max_inter* |
| Step 4 | the number of high-frequency patterns to select as columns for the matrix | 4,000 | *num_patterns* |
| Step 10 | the number of dimensions in the projection | 300 | *k* |
| Step 11 | the number of ways to compare a version of *A:B* with a version of *C:D* | 16 | *num_combinations* |

# 5 Experiments with Word Analogy Questions

This section presents various experiments with 374 multiple-choice SAT word analogy questions (Turney and Littman, 2003b, 2005). Section 5.1 evaluates LRA, exactly as described in the previous section, with the 374 questions. Section 5.2 compares the performance of LRA to related work and Section 5.3 looks at human performance on the questions. Section 5.4 examines the effect of varying *k*, the number of dimensions for the SVD projection (step 10 in Section 4). The remaining subsections perform ablation experiments, exploring the consequences of removing some steps from LRA.

## 5.1 Baseline LRA System

LRA correctly answered 210 of the 374 questions. 160 questions were answered incorrectly and 4 questions were skipped, because the stem pair and its alternates were represented by zero vectors. For example, one of the skipped questions had the stem heckler:disconcert (solution: heckler:disconcert::lobbyist:persuade). These two words did not appear together in the WMTS corpus, in any phrase of five words or less, so they were dropped in step 5 (Section 4), since they would be represented by a zero vector. Furthermore, none of the 20 alternate pairs (from step 1) appeared together in any phrase of five words or less.

Since there are five choices for each question, we would expect to answer 20% of the questions correctly by random guessing. Following Turney *et al.* (2003), we score the performance by giving one point for each correct answer and 0.2 points for each skipped question, so LRA attained a score of 56.4% on the 374 SAT questions.

With 374 questions and 6 word pairs per question (one stem and five choices), there are 2,244 pairs in the input set. In step 2, introducing alternate pairs multiplies the number of pairs by four, resulting in 8,976 pairs. In step 5, for each pair *A:B,* we add *B:A,* yielding 17,952 pairs. However, some pairs are dropped because they correspond to zero vectors (they do not appear together in a window of five words in the WMTS corpus). Also, a few words do not appear in Lin's thesaurus, and some word pairs appear twice in the SAT questions (e.g., lion:cat). The sparse matrix (step 7) has 17,232 rows (word pairs) and 8,000 columns (patterns), with a density of 5.8% (percentage of nonzero values).

Table 9 gives the time required for each step of LRA, a total of almost nine days. All of the steps used a single CPU on a desktop computer, except step 3, finding the phrases for each word pair, which used a 16 CPU Beowulf cluster. Most of the other steps are parallelizable; with a bit of programming effort, they could also be executed on the Beowulf cluster. All CPUs (both desktop and cluster) were 2.4 GHz Intel Xeons. The desktop computer had 2 GB of RAM and the cluster had a total of 16 GB of RAM.

Table 9. LRA elapsed run time.

| Step | Description | Time H:M:S | Hardware |
|------|-------------|------------|----------|
| 1 | Find alternates | 24:56:00 | 1 CPU |
| 2 | Filter alternates | 0:00:02 | 1 CPU |
| 3 | Find phrases | 109:52:00 | 16 CPUs |
| 4 | Find patterns | 33:41:00 | 1 CPU |
| 5 | Map pairs to rows | 0:00:02 | 1 CPU |
| 6 | Map patterns to columns | 0:00:02 | 1 CPU |
| 7 | Generate a sparse matrix | 38:07:00 | 1 CPU |
| 8 | Calculate entropy | 0:11:00 | 1 CPU |
| 9 | Apply SVD | 0:43:28 | 1 CPU |
| 10 | Projection | 0:08:00 | 1 CPU |
| 11 | Evaluate alternates | 2:11:00 | 1 CPU |
| 12 | Calculate relational similarity | 0:00:02 | 1 CPU |
| Total | | 209:49:36 | |

## 5.2    LRA versus VSM and PRMC

Table 10 compares LRA to the Vector Space Model with the 374 analogy questions. VSM-AV refers to the VSM using AltaVista's database as a corpus. The VSM-AV results are taken from Turney and Littman (2003b, 2005). As mentioned in Section 3, we estimate this corpus contained about $5 \times 10^{11}$ English words at the time the VSM-AV experiments took place. Turney and Littman (2003b, 2005) gave an estimate of $1 \times 10^{11}$ English words, but we believe this estimate was slightly conservative. VSM-WMTS refers to the VSM using the WMTS, which contains about 5 x $10^{10}$ English words.[9] We generated the VSM-WMTS results by adapting the VSM to the WMTS. The algorithm is slightly different from Turney and Littman (2003b, 2005), because we used passage frequencies instead of document frequencies.

Table 10. LRA versus VSM with 374 SAT analogy questions.

|  | VSM-AV | VSM-WMTS | LRA |
|---|---|---|---|
| Correct | 176 | 144 | 210 |
| Incorrect | 193 | 196 | 160 |
| Skipped | 5 | 34 | 4 |
| Total | 374 | 374 | 374 |
| Score | 47.3% | 40.3% | 56.4% |

All three pairwise differences in the three scores in Table 10 are statistically significant with 95% confidence, using the Fisher Exact Test. Using the same corpus as the VSM, LRA achieves a score of 56% whereas the VSM achieves a score of 40%, an absolute difference of 16% and a relative improvement of 40%. When VSM has a corpus ten times larger than LRA's corpus, LRA is still ahead, with an absolute difference of 9% and a relative improvement of 19%.

Comparing VSM-AV to VSM-WMTS, the smaller corpus has reduced the score of the VSM, but much of the drop is due to the larger number of questions that were skipped (34 for VSM-WMTS versus 5 for VSM-AV). With the smaller corpus, many more of the input word pairs simply do not appear together in short phrases in the corpus. LRA is able to answer as many questions as VSM-AV, although it uses the same corpus as VSM-WMTS, because Lin's thesaurus allows LRA to substitute synonyms for words that are not in the corpus.

VSM-AV required 17 days to process the 374 analogy questions (Turney and Littman, 2003b, 2005), compared to 9 days for LRA. As a courtesy to AltaVista, Turney and Littman (2003b, 2005) inserted a five second delay between each query. Since the WMTS is running locally, there is no need for delays. VSM-WMTS processed the questions in only one day.

Table 11 compares LRA to Product Rule Module Combination (Turney *et al.*, 2003). The results for PRMC are taken from Turney *et al.* (2003). Since PRMC is a supervised algorithm, 274 of the analogy questions are used for training and 100 are used for testing. Results are given for two different splits of the questions.

Table 11. LRA versus PRMC with 100 SAT questions.

|  | Testing subset #1 | | Testing subset #2 | |
|---|---|---|---|---|
|  | PRMC | LRA | PRMC | LRA |
| Correct | 45 | 51 | 55 | 59 |
| Incorrect | 55 | 49 | 45 | 39 |
| Skipped | 0 | 0 | 0 | 2 |
| Total | 100 | 100 | 100 | 100 |
| Score | 45.0% | 51.0% | 55.0% | 59.4% |

Due to the relatively small sample size (100 questions instead of 374), the differences between LRA and PRMC are not statistically significant with 95% confidence. However,

---

[9] The Waterloo Terabyte Corpus contains exactly 55,156,882,530 words, but not all are English.

LRA appears to have higher scores than PRMC, although LRA is not supervised. As mentioned in Section 3, VSM-AV is one of the thirteen modules in PRMC.

## 5.3    Human Performance

The average performance of college-bound senior high school students on verbal SAT questions corresponds to a score of about 57% (Turney and Littman, 2003b, 2005). Table 12 gives the 95% confidence intervals for LRA, VSM, and PRMC, calculated by the Binomial Exact Test. In all cases, there is no significant difference between LRA and human performance. VSM-AV and VSM-WMTS are significantly below human-level performance. PRMC appears to be within human levels with one testing subset, but not with the other. However, these subsets have very wide 95% confidence intervals, due to the relatively small sample size.

Table 12. Comparison with human SAT performance.

| System | Questions | Score | 95% confidence interval | | Human-level 57% |
|---|---|---|---|---|---|
| | | | Lower | Upper | |
| VSM-AV | all 374 | 47.3% | 42.2% | 52.5% | NO |
| VSM-WMTS | all 374 | 40.3% | 35.4% | 45.5% | NO |
| LRA | all 374 | 56.4% | 51.2% | 61.5% | YES |
| PRMC | 100 subset #1 | 45.0% | 35.0% | 55.3% | NO |
| LRA | 100 subset #1 | 51.0% | 40.8% | 61.1% | YES |
| PRMC | 100 subset #2 | 55.0% | 45.7% | 65.9% | YES |
| LRA | 100 subset #2 | 59.4% | 49.7% | 69.7% | YES |

## 5.4    Varying the Number of Dimensions in SVD

Table 13 shows the variation in the score of LRA as the number of dimensions in the SVD projection (step 10 in Section 4) varies from 100 to 1000. We use $k = 300$ because Landauer and Dumais (1997) found that this value worked well when measuring attributional similarity between words. The difference between the lowest score (55.3%) and the highest score (57.2%) is not statistically significant with 95% confidence, according to the Fisher Exact Test. It seems that LRA is not particularly sensitive to the value of $k$.

Table 13. LRA's score as a function of the number of dimensions *(k)*.

| $k$ | Correct | Incorrect | Skipped | Total | Score |
|---|---|---|---|---|---|
| 100 | 206 | 164 | 4 | 374 | 55.3% |
| 200 | 209 | 161 | 4 | 374 | 56.1% |
| **300** | **210** | **160** | **4** | **374** | **56.4%** |
| 400 | 209 | 161 | 4 | 374 | 56.1% |
| 500 | 213 | 157 | 4 | 374 | 57.2% |
| 600 | 209 | 161 | 4 | 374 | 56.1% |
| 700 | 209 | 161 | 4 | 374 | 56.1% |
| 800 | 210 | 160 | 4 | 374 | 56.4% |
| 900 | 208 | 162 | 4 | 374 | 55.8% |
| 1000 | 208 | 162 | 4 | 374 | 55.8% |

## 5.5    LRA without SVD

With no SVD, LRA's score drops from 56.4% to 53.2%. This drop (3.2%) is not statistically significant with 95% confidence, according to the Fisher Exact Test. However, it is worth noting that dropping SVD *increases* the execution time of LRA by 11 hours. Although dropping SVD eliminates steps 9 and 10, and thus saves almost one hour (see Table 9), the execution time for step 11 increases by almost 12 hours, because we are now calculating the cosines between 8,000 dimensional row vectors, instead of 300 dimensional row vectors.

### 5.6    LRA without Synonyms

Without synonyms (step 1 in Section 4), LRA's score drops from 56.4% to 50.6%. The decrease (5.8%) is not statistically significant with 95% confidence. It is interesting that the number of skipped questions rises from 4 to 22, which demonstrates the value of synonyms for compensating for sparse data.

### 5.7    LRA without Synonyms and without SVD

If we eliminate both synonyms and SVD from LRA, all that distinguishes LRA from VSM-WMTS is the patterns (step 4). The VSM approach uses a fixed list of 64 patterns to generate 128 dimensional vectors (Turney and Littman, 2003b, 2005), whereas LRA uses a dynamically generated set of 4,000 patterns, resulting in 8,000 dimensional vectors. Table 14 compares three systems, VSM-WMTS, LRA with neither SVD nor synonyms, and baseline LRA. All three pairwise differences in the three scores are statistically significant with 95% confidence, using the Fisher Exact Test. Although the benefit of SVD and synonyms is not significant when they are considered independently, their combined benefit is statistically significant. The table also shows that the 4,000 automatically generated patterns are superior to the 64 hand-generated patterns.

Table 14. LRA with neither synonyms nor SVD.

|  | VSM-WMTS | LRA no SVD no synonyms | LRA baseline system |
|---|---|---|---|
| Correct | 144 | 178 | 210 |
| Incorrect | 196 | 173 | 160 |
| Skipped | 34 | 23 | 4 |
| Total | 374 | 374 | 374 |
| Score | 40.3% | 48.8% | 56.4% |

Moving from 64 patterns (VSM-WMTS) to 4,000 patterns enables us to handle eleven questions that were previously skipped (compare the second and third columns in Table 14; from 34 skipped to 23 skipped). Adding SVD allows us to handle one more skipped question (see Section 5.6; from 23 skipped to 22 skipped). Finally, adding synonyms lets us handle eighteen more questions (see the fourth column in Table 14; from 22 skipped to 4 skipped). Recall has increased from 38.5% for VSM-WMTS (144/374) to 56.1% for LRA (210/374), and precision has also increased from 42.4% (144/340) to 56.8% (210/370).

## 6   Experiments with Noun-Modifier Relations

This section describes experiments with 600 noun-modifier pairs, hand-labeled with 30 classes of semantic relations (Nastase and Szpakowicz, 2003). Section 6.1 evaluates LRA as an unsupervised component in a simple single nearest neighbour supervised learning algorithm. LRA serves as the distance measure (equivalently, nearness measure) for determining which training word pair is the nearest neighbour for a given testing word pair. Section 6.2 compares the performance LRA to VSM, when they are used as distance measures for classifying semantic relations.

## 6.1 Baseline LRA with Single Nearest Neighbour

We experiment with both a 30 class problem and a 5 class problem. The 30 classes of semantic relations include *cause* (e.g., in "flu virus", the head noun "virus" is the *cause* of the modifier "flu"), *location* (e.g., in "home town", the head noun "town" is the *location* of the modifier "home"), *part* (e.g., in "printer tray", the head noun "tray" is *part* of the modifier "printer"), and *topic* (e.g., in "weather report", the head noun "report" is about the *topic* "weather"). For a full list of classes, see Nastase and Szpakowicz (2003) or Turney and Littman (2003b, 2005). The 30 classes belong to 5 general groups of relations, *causal* relations, *temporal* relations, *spatial* relations, *participatory* relations (e.g., in "student protest", the "student" is the *agent* who performs the "protest"; *agent* is a *participatory* relation), and *qualitative* relations (e.g., in "oak tree", "oak" is a *type* of "tree"; *type* is a *qualitative* relation).

The following experiments use single nearest neighbour classification with leave-one-out cross-validation. For leave-one-out cross-validation, the testing set consists of a single noun-modifier pair and the training set consists of the 599 remaining noun-modifiers. The data set is split 600 times, so that each noun-modifier gets a turn as the testing word pair. The predicted class of the testing pair is the class of the single nearest neighbour in the training set. As the measure of nearness, we use LRA to calculate the relational similarity between the testing pair and the training pairs.

Each SAT question has five choices, so answering 374 SAT questions required calculating $374 \times 5 \times 16 = 29,920$ cosines. The factor of 16 comes from the alternate pairs, step 11 in LRA. With the noun-modifier pairs, using leave-one-out cross-validation, each test pair has 599 choices, so an exhaustive application of LRA would require calculating $600 \times 599 \times 16 = 5,750,400$ cosines. To reduce the amount of computation required, we first find the 30 nearest neighbours for each pair, ignoring the alternate pairs ($600 \times 599 = 359,400$ cosines), and then apply the full LRA, including the alternates, to just those 30 neighbours ($600 \times 30 \times 16 = 288,000$ cosines), which requires calculating only $359,400 + 288,000 = 647,400$ cosines.

There are 600 word pairs in the input set for LRA. In step 2, introducing alternate pairs multiplies the number of pairs by four, resulting in 2,400 pairs. In step 5, for each pair *A:B,* we add *B:A,* yielding 4,800 pairs. However, some pairs are dropped because they correspond to zero vectors and a few words do not appear in Lin's thesaurus. The sparse matrix (step 7) has 4,748 rows and 8,000 columns, with a density of 8.4%.

Following Turney and Littman (2003b, 2005), we evaluate the performance by accuracy and also by the macroaveraged F measure (Lewis, 1991). The F measure is the harmonic mean of precision and recall. Macroaveraging calculates the precision, recall, and F for each class separately, and then calculates the average across all classes. On the 30 class problem, LRA with the single nearest neighbour algorithm achieves an accuracy of 39.8% (239/600) and a macroaveraged F of 36.6%. On the 5 class problem, the accuracy is 58.0% (348/600) and the macroaveraged F is 54.6%.

## 6.2 LRA versus VSM

Table 15 shows the performance of LRA and VSM on the 30 class problem. VSM-AV is VSM with the AltaVista corpus and VSM-WMTS is VSM with the WMTS corpus. The results for VSM-AV are taken from Turney and Littman (2003b, 2005). All three pairwise differences in the three F measures are statistically significant at the 95% level, according to the Paired T-Test. The accuracy of LRA is significantly higher than the accuracies of VSM-AV and VSM-WMTS, according to the Fisher Exact Test, but the difference between the two VSM accuracies is not significant. Using the same corpus as the VSM, LRA's accuracy is 15% higher in absolute terms (39.8% − 24.7%) and 61% higher in relative terms (39.8% / 24.7%).

Table 15. Comparison of LRA and VSM on the 30 class problem.

|           | VSM-AV | VSM-WMTS | LRA   |
|-----------|--------|----------|-------|
| Correct   | 167    | 148      | 239   |
| Incorrect | 433    | 452      | 361   |
| Total     | 600    | 600      | 600   |
| Accuracy  | 27.8%  | 24.7%    | 39.8% |
| Precision | 27.9%  | 24.0%    | 41.0% |
| Recall    | 26.8%  | 20.9%    | 35.9% |
| F         | 26.5%  | 20.3%    | 36.6% |

Table 16 compares the performance of LRA and VSM on the 5 class problem. The accuracy and F measure of LRA are significantly higher than the accuracies and F measures of VSM-AV and VSM-WMTS, but the differences between the two VSM accuracies and F measures are not significant. Using the same corpus as the VSM, LRA's accuracy is 14% higher in absolute terms (58.0% − 44.0%) and 32% higher in relative terms (58.0% / 44.0%).

Table 16. Comparison of LRA and VSM on the 5 class problem.

|           | VSM-AV | VSM-WMTS | LRA   |
|-----------|--------|----------|-------|
| Correct   | 274    | 264      | 348   |
| Incorrect | 326    | 336      | 252   |
| Total     | 600    | 600      | 600   |
| Accuracy  | 45.7%  | 44.0%    | 58.0% |
| Precision | 43.4%  | 40.2%    | 55.9% |
| Recall    | 43.1%  | 41.4%    | 53.6% |
| F         | 43.2%  | 40.6%    | 54.6% |

## 7  Discussion

The experimental results in Sections 5 and 6 demonstrate that LRA performs significantly better than the VSM, but it is also clear that there is room for improvement. The accuracy might not yet be adequate for the practical applications mentioned in Section 2, although past work has shown that it is possible to adjust the tradeoff of precision versus recall (Turney and Littman, 2003b, 2005). For some of the applications, such as information extraction, LRA might be suitable if it is adjusted for high precision, at the expense of low recall.

Another limitation is speed; it took almost nine days for LRA to answer 374 analogy questions. However, with progress in computer hardware, speed will gradually become less of a concern. Also, the software has not been optimized for speed; there are several places where the efficiency could be increased and many operations are parallelizable. It may also be possible to precompute much of the information for LRA, although this would require substantial changes to the algorithm.

The difference in performance between VSM-AV and VSM-WMTS shows that VSM is sensitive to the size of the corpus. Although LRA is able to surpass VSM-AV when the WMTS corpus is only about one tenth the size of the AV corpus, it seems likely that LRA would perform better with a larger corpus. The WMTS corpus requires one terabyte of hard disk space, but progress in hardware will likely make ten or even one hundred terabytes affordable in the relatively near future.

For noun-modifier classification, more labeled data should yield performance improvements. With 600 noun-modifier pairs and 30 classes, the average class has only 20 examples. We expect that the accuracy would improve substantially with five or ten times more examples. Unfortunately, it is time consuming and expensive to acquire hand-labeled data.

Another issue with noun-modifier classification is the choice of classification scheme for the semantic relations. The 30 classes of Nastase and Szpakowicz (2003) might not be the best scheme. Other researchers have proposed different schemes (Vanderwende, 1994; Barker and Szpakowicz, 1998; Rosario and Hearst, 2001; Rosario *et al.*, 2002). It seems likely that some schemes are easier for machine learning than others. For some of the applications we consider (Section 2), 30 classes may not be necessary; the 5 class scheme may be sufficient.

There are several numerical parameters in LRA, listed in Table 8. So far, we have only experimentally evaluated the number of dimensions, *k* (Section 5.4). The remaining parameters were set at arbitrary values. In future work, we will investigate the influence of these parameters on the performance. We hypothesize that LRA will be relatively insensitive to the exact values of the parameters.

LRA, like VSM, is a corpus-based approach to measuring relational similarity. Past work suggests that a hybrid approach, combining multiple modules, some corpus-based, some lexicon-based, will surpass any purebred approach (Turney *et al.,* 2003). Veale's (2003) work hints at a lexicon-based approach to measuring relational similarity and he has had some promising provisional results (personal communication). We believe a good research strategy is to push the performance of each purebred approach as far as possible, before combining them in a hybrid approach.

The Singular Value Decomposition is only one of many methods for handling sparse, noisy data. We have also experimented with Nonnegative Matrix Factorization (NMF) (Lee and Seung, 1999), Probabilistic Latent Semantic Analysis (PLSA) (Hofmann, 1999), Kernel Principal Components Analysis (KPCA) (Scholkopf *et al.,* 1997), and Iterative Scaling (IS) (Ando, 2000). We had some interesting results with small matrices (around 2,000 rows by 1,000 columns), but none of these methods seemed substantially better than SVD and none of them scaled up to the matrix sizes we are using here (e.g., 17,232 rows and 8,000 columns; see Section 5.1).

# 8 Conclusion

This paper has introduced a new method for calculating relational similarity, Latent Relational Analysis. The experiments demonstrate that LRA performs better than the VSM approach, when evaluated with SAT word analogy questions and with the task of classifying noun-modifier expressions. The VSM approach represents the relation between a pair of words with a vector, in which the elements are based on the frequencies of 64 hand-built patterns in a large corpus. LRA extends this approach in three ways: (1) the patterns are generated dynamically from the corpus, (2) SVD is used to smooth the data, and (3) a thesaurus is used to explore reformulations of the word pairs. Ablation experiments show that each of these three extensions contributes to the improved performance of LRA over the VSM.

We have presented several examples of the many potential applications for measures of relational similarity. Just as attributional similarity measures have proven to have many practical uses, we expect that relational similarity measures will soon become widely used. Gentner *et al.* (2001) argue that relational similarity is essential to understanding novel metaphors (as opposed to conventional metaphors). Many researchers have argued that metaphor is the heart of human thinking (Lakoff and Johnson, 1980; Lakoff, 1987; Hofstadter *et al.*, 1995; French, 2002). We believe that relational similarity plays a fundamental role in the mind and therefore relational similarity measures could be crucial for artificial intelligence.

In future work, we plan to investigate some potential applications for LRA. It is possible that the error rate of LRA is still too high for practical applications, but the fact that LRA matches average human performance on SAT analogy questions is encouraging.

## Acknowledgements

Thanks to Michael Littman for sharing the 374 SAT analogy questions and for inspiring me to tackle them. Thanks to Vivi Nastase and Stan Szpakowicz for sharing their 600 classified noun-modifier phrases. Thanks to Egidio Terra, Charlie Clarke, and the School of Computer Science of the University of Waterloo, for giving us a copy of the Waterloo MultiText System and their Terabyte Corpus. Thanks to Dekang Lin for making his Dependency-Based Word Similarity lexicon available online. Thanks to Doug Rohde for SVDLIBC and Michael Berry for SVDPACK.

## References

Ando, R.K. (2000). Latent semantic space: Iterative scaling improves inter-document similarity measurement. *Proceedings of the 23rd Annual International ACM SIGIR*, 216-223.

Baeza-Yates, R., and Ribeiro-Neto, B. (1999). *Modern Information Retrieval*. Addison-Wesley.

Banerjee, S., and Pedersen, T. (2003). Extended gloss overlaps as a measure of semantic relatedness. *Proceedings of the Eighteenth International Joint Conference on Artificial Intelligence (IJCAI-03)*. Acapulco, Mexico, 805-810.

Barker, K., and Szpakowicz, S. (1998). Semi-automatic recognition of noun modifier relationships. *Proceedings of the 17th International Conference on Computational Linguistics and the 36th Annual Meeting of the Association for Computational Linguistics (COLING-ACL'98)*, Montréal, Québec, 96-102.

Barzilay, R., and Elhadad, M. (1997). Using lexical chains for text summarization. *Proceedings of the ACL'97/EACL'97 Workshop on Intelligent Scalable Text Summarization*, 10-17.

Berland, M. and Charniak, E. (1999). Finding parts in very large corpora. *Proceedings of the 37th Annual Meeting of the Association for Computational Linguistics (ACL '99)*. ACL, New Brunswick NJ, 57-64.

Berry, M.W. (1992). Large scale singular value computations. *International Journal of Supercomputer Applications*, 6(1), 13-49.

Budanitsky, A., and Hirst, G. (2001). Semantic distance in WordNet: An experimental, application-oriented evaluation of five measures. *Proceedings of the Workshop on WordNet and Other Lexical Resources, Second Meeting of the North American Chapter of the Association for Computational Linguistics,* Pittsburgh, 29-34.

Church, K.W., And Hanks, P. (1989). Word association norms, mutual information and lexicography. *Proceedings of the 27th Annual Conference of the Association of Computational Linguistics.* Association for Computational Linguistics, New Brunswick, NJ, 76-83.

Claman, C. (2000). *10 Real SATs*. College Entrance Examination Board.

Clarke, C.L.A., Cormack, G.V., and Palmer, C.R. (1998). An overview of MultiText. *ACM SIGIR Forum,* 32(2), 14-15.

Daganzo, C.F. (1994). The cell transmission model: A dynamic representation of highway traffic consistent with the hydrodynamic theory. *Transportation Research Part B: Methodological*, 28(4), 269-287.

Deerwester, S., Dumais, S.T., Furnas, G.W., Landauer, T.K., and Harshman, R. (1990). Indexing by latent semantic indexing. *Journal of the American Society for Information Science (JASIS),* 41(6), 391-407.

Dolan, W.B. (1995). Metaphor as an emergent property of machine-readable dictionaries. *Proceedings of the AAAI 1995 Spring Symposium Series: Representation and Acquisition of Lexical Knowledge: Polysemy, Ambiguity and Generativity,* 27-32.

Dumais, S.T. (1990). *Enhancing Performance in Latent Semantic Indexing (LSI) Retrieval.* TM-ARH-017527 Technical Report, Bellcore.

Dumais, S.T. (1993). Latent semantic indexing (LSI) and TREC-2. *Proceedings of the Second Text REtrieval Conference (TREC-2)*, D.K. Harman, Ed., National Institute of Standards and Technology, 105-115.

Dunning, T. (1993). Accurate methods for the statistics of surprise and coincidence. *Computational Linguistics*, 19, 61-74.

Falkenhainer, B., Forbus, K.D., and Gentner, D. (1989). The structure-mapping engine: Algorithm and examples. *Artificial Intelligence*, 41(1), 1-63.

Falkenhainer, B. (1990). Analogical interpretation in context. *Proceedings of the Twelfth Annual Conference of the Cognitive Science Society*, Lawrence Erlbaum Associates, 69-76.

Fellbaum, C. (editor). (1998). *WordNet: An Electronic Lexical Database*. MIT Press.

Foltz, P.W., Kintsch, W., and Landauer, T.K. (1998). The measurement of textual coherence with latent semantic analysis. *Discourse Processes,* 25, 285-307.

Foltz, P.W., Laham, D., and Landauer, T.K. (1999). Automated essay scoring: Applications to educational technology. *Proceedings of the ED-MEDIA '99 Conference,* Association for the Advancement of Computing in Education, Charlottesville.

French, R.M. (2002). The computational modeling of analogy-making. *Trends in Cognitive Sciences*, 6(5), 200-205.

Gentner, D. (1983). Structure-mapping: A theoretical framework for analogy. *Cognitive Science*, 7(2), 155-170.

Gentner, D., Bowdle, B., Wolff, P., and Boronat, C. (2001). Metaphor is like analogy. In D. Gentner, K.J. Holyoak, and B. Kokinov (Eds.), *The Analogical Mind: Perspectives from Cognitive Science.* Cambridge, MA: MIT Press.

Gildea, D., Jurafsky, D. (2002). Automatic labeling of semantic roles. *Computational Linguistics*, 28(3), 245-288.

Golub, G.H., and Van Loan, C.F. (1996). *Matrix Computations.* Third edition. Johns Hopkins University Press, Baltimore, MD.

Harman, D. (1986). An experimental study of factors important in document ranking. *Proceedings of the Ninth Annual International ACM SIGIR Conference on Research and Development in Information Retrieval (SIGIR'86).* Pisa, Italy. 186-193.

Hearst, M.A. (1992a). Automatic acquisition of hyponyms from large text corpora. *Proceedings of the Fourteenth International Conference on Computational Linguistics*, Nantes, France, 539-545.

Hearst, M.A. (1992b). Direction-based text interpretation as an information access refinement. In P. Jacobs (Ed.), *Text-Based Intelligent Systems: Current Research and Practice in Information Extraction and Retrieval.* Mahwah, NJ: Lawrence Erlbaum Associates.

Hofmann, T. (1999). Probabilistic latent semantic indexing. *Proceedings of the 22$^{nd}$ Annual ACM Conference on Research and Development in Information Retrieval (SIGIR '99),* Berkeley, California, 50-57.

Hofstadter, D., and the Fluid Analogies Research Group (1995). *Fluid Concepts and Creative Analogies: Computer Models of the Fundamental Mechanisms of Thought.* New York: Basic Books.

Jarmasz, M. and Szpakowicz, S. (2003). Roget's thesaurus and semantic similarity. *Proceedings of the International Conference on Recent Advances in Natural Language Processing (RANLP-03),* Borovets, Bulgaria, 212-219.

Jiang, J., and Conrath, D. (1997). Semantic similarity based on corpus statistics and lexical taxonomy. *Proceedings of the International Conference on Research in Computational Linguistics (ROCLING X)*, Tapei, Taiwan,19-33.

Lakoff, G., and Johnson, M. (1980). *Metaphors We Live By*. University of Chicago Press.

Lakoff, G. (1987). *Women, Fire, and Dangerous Things.* University of Chicago Press.

Landauer, T.K., and Dumais, S.T. (1997). A solution to Plato's problem: The latent semantic analysis theory of the acquisition, induction, and representation of knowledge. *Psychological Review,* 104, 211-240.

Lee, D.D., Seung, H.S. (1999). Learning the parts of objects by nonnegative matrix factorization. *Nature*, 401, 788-791.

Lesk, M.E. (1969). Word-word associations in document retrieval systems. *American Documentation*, 20(1): 27-38.

Lesk, M.E. (1986). Automatic sense disambiguation using machine readable dictionaries: How to tell a pine cone from a ice cream cone. *Proceedings of ACM SIGDOC '86*, 24-26.

Lewis, D.D. (1991). Evaluating text categorization. *Proceedings of the Speech and Natural Language Workshop,* Asilomar, 312-318.

Lin, D. (1998a). An information-theoretic definition of similarity. *Proceedings of the Fifteenth International Conference on Machine Learning (ICML '98),* 296-304.

Lin, D. (1998b). Automatic retrieval and clustering of similar words. *Proceedings of the 36th Annual Meeting of the Association for Computational Linguistics and the 17th International Conference on Computational Linguistics (COLING-ACL '98)*, Montreal, Canada, 768-774.

Lin, D. (1998c). Dependency-based evaluation of MINIPAR. *Proceedings of the Workshop at LREC'98 (First International Conference on Language Resources and Evaluation) on the Evaluation of Parsing Systems*, Granada, Spain.

Martin, J. (1992). Computer understanding of conventional metaphoric language. *Cognitive Science*, 16, 233-270.

Medin, D.L., Goldstone, R.L., and Gentner, D. (1990). Similarity involving attributes and relations: Judgments of similarity and difference are not inverses. *Psychological Science,* 1(1), 64-69.

Medin, D.L., Goldstone, R.L., and Gentner, D. (1993). Respects for similarity. *Psychological Review,* 100(2), 254-278.

Mel'cuk, I. (1988). *Dependency Syntax: Theory and Practice.* New York: State University of New York Press.

Morris, J., and Hirst, G. (1991). Lexical cohesion computed by thesaural relations as an indicator of the structure of text. *Computational Linguistics,* 17(1), 21-48.

Nastase, V., and Szpakowicz, S. (2003). Exploring noun-modifier semantic relations. *Fifth International Workshop on Computational Semantics (IWCS-5)*, Tilburg, The Netherlands, 285-301.

Paice, C.D., and Black, W.J. (2003). A three-pronged approach to the extraction of key terms and semantic roles. *Proceedings of the International Conference on Recent Advances in Natural Language Processing (RANLP-03),* Borovets, Bulgaria. 357-363.

Pantel, P., and Lin, D. (2002). Discovering word senses from text. *Proceedings of ACM SIGKDD Conference on Knowledge Discovery and Data Mining,* 613-619.

Patwardhan, S., Banerjee, S., and Pedersen, T. (2003). Using measures of semantic relatedness for word sense disambiguation. *Proceedings of the Fourth International Conference on Intelligent Text Processing and Computational Linguistics*, Mexico City.

Rehder, B., Schreiner, M.E., Wolfe, M.B., Laham, D., Landauer, T.K., and Kintsch, W. (1998). Using latent semantic analysis to assess knowledge: Some technical considerations. *Discourse Processes,* 25, 337-354.

Reitman, W.R. (1965). *Cognition and Thought: An Information Processing Approach.* New York, NY: John Wiley and Sons.

Resnik, P. (1995). Using information content to evaluate semantic similarity in a taxonomy. *Proceedings of the 14th International Joint Conference on Artificial Intelligence.* Morgan Kaufmann, San Mateo, CA, 448-453.

Rosario, B., and Hearst, M. (2001). Classifying the semantic relations in noun-compounds via a domain-specific lexical hierarchy. *Proceedings of the 2001 Conference on Empirical Methods in Natural Language Processing (EMNLP-01)*, 82-90.

Rosario, B, Hearst, M., and Fillmore, C. (2002). The descent of hierarchy, and selection in relational semantics. *Proceedings of the 40th Annual Meeting of the Association for Computational Linguistics (ACL '02)*, Philadelphia, PA, 417-424.

Ruge, G. (1992). Experiments on linguistically-based term associations. *Information Processing and Management*, 28(3), 317-332.

Ruge, G. (1997). Automatic detection of thesaurus relations for information retrieval applications. *Foundations of Computer Science: Potential - Theory - Cognition*, C. Freksa, M. Jantzen, R. Valk (Eds.), Lecture Notes in Computer Science, Springer-Verlag, 499-506.

Salton, G., and McGill, M.J. (1983). *Introduction to Modern Information Retrieval.* McGraw-Hill, New York.

Salton, G. (1989). *Automatic Text Processing: The Transformation, Analysis, and Retrieval of Information by Computer.* Addison-Wesley, Reading, Massachusetts.

Salton, G., and Buckley, C. (1988). Term-weighting approaches in automatic text retrieval. *Information Processing and Management*, 24(5), 513-523.

Scholkopf, B., Smola, A.J., and Muller, K. (1997). Kernel principal component analysis. *Proceedings of the International Conference on Artificial Neural Networks (ICANN-1997)*, Berlin, 583-588.

Smadja, F. (1993). Retrieving collocations from Text: Xtract. *Computational Linguistics*, 19, 143-177.

Terra, E., and Clarke, C.L.A. (2003). Frequency estimates for statistical word similarity measures. *Proceedings of the Human Language Technology and North American Chapter of Association of Computational Linguistics Conference 2003 (HLT/NAACL 2003)*, 244–251.

Turney, P.D. (2001). Mining the Web for synonyms: PMI-IR versus LSA on TOEFL. *Proceedings of the Twelfth European Conference on Machine Learning.* Springer-Verlag, Berlin, 491-502.

Turney, P.D. (2002). Thumbs up or thumbs down? Semantic orientation applied to unsupervised classification of reviews. *Proceedings of the 40th Annual Meeting of the Association for Computational Linguistics (ACL'02).* Philadelphia, Pennsylvania, 417-424.

Turney, P.D. (2003). Coherent keyphrase extraction via Web mining. *Proceedings of the Eighteenth International Joint Conference on Artificial Intelligence (IJCAI-03).* Acapulco, Mexico, 434-439.

Turney, P.D. (2004). Word sense disambiguation by Web mining for word co-occurrence probabilities. *Proceedings of the Third International Workshop on the Evaluation of*

*Systems for the Semantic Analysis of Text (SENSEVAL-3),* Barcelona, Spain, 239-242.

Turney, P.D., Littman, M.L., Bigham, J., and Shnayder, V. (2003). Combining independent modules to solve multiple-choice synonym and analogy problems. *Proceedings of the International Conference on Recent Advances in Natural Language Processing (RANLP-03).* Borovets, Bulgaria, 482-489.

Turney, P.D., and Littman, M.L. (2003a). Measuring praise and criticism: Inference of semantic orientation from association. *ACM Transactions on Information Systems (TOIS)*, 21 (4), 315-346.

Turney, P.D., and Littman, M.L. (2003b). *Learning Analogies and Semantic Relations*, National Research Council, Institute for Information Technology, Technical Report ERB-1103.

Turney, P.D., and Littman, M.L. (2005). Corpus-based learning of analogies and semantic relations. *Machine Learning,* in press.

Vanderwende, L. (1994). Algorithm for automatic interpretation of noun sequences. *Proceedings of the Fifteenth International Conference on Computational Linguistics*, Kyoto, Japan, 782-788.

Veale, T. (2003). The analogical thesaurus. *Proceedings of the 15th Innovative Applications of Artificial Intelligence Conference (IAAI 2003)*, Acapulco, Mexico, 137-142.

Yarowsky, D. (1993). One sense per collocation. *Proceedings of the ARPA Human Language Technology Workshop*. Princeton, 266-271.

Yarowsky, D. (1995). Unsupervised word sense disambiguation rivaling supervised methods. *Proceedings of the 33rd Annual Meeting of the Association for Computational Linguistics*. Cambridge, MA, 189-196.

Yi, J., Lin, H., Alvarez, L., and Horowitz, R. (2003). Stability of macroscopic traffic flow modeling through wavefront expansion. *Transportation Research Part B: Methodological*, 37(7), 661-679.

Zhang, H.M. (2003). Driver memory, traffic viscosity and a viscous vehicular traffic flow model. *Transportation Research Part B: Methodological*, 37(1), 27-41.